%% file: main.tex
\documentclass[10pt,twocolumn,letterpaper]{article}

\usepackage{wacv}              
\usepackage{algorithm}
\usepackage[accsupp]{axessibility} 
\usepackage{algorithmic}
\usepackage{makecell}
\usepackage{bbm}
\usepackage{subcaption}
\usepackage{pifont}
\usepackage{booktabs}

\definecolor{wacvblue}{rgb}{0.21,0.49,0.74}
\usepackage[pagebackref,breaklinks,colorlinks,allcolors=wacvblue]{hyperref}

\def\wacvPaperID{775} 
\def\confName{WACV}
\def\confYear{2026}
\usepackage{multirow}
\newcommand{\clipit}{\texttt{CLIP-IT}\xspace}

\DeclareRobustCommand{\logo}{%
  \begingroup\normalfont
  \raisebox{-0.25em}{%
  \hspace{-0.5em}
  \includegraphics[height=3.0em]{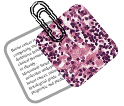}%
  }%
  \kern 0.2em
  \endgroup
}

\title{\logo \clipit: \underline{CLIP}-based Pairing of Histology \underline{I}mages  with Privileged \underline{T}extual Information}
\author{

Banafsheh Karimian$^{1}$
\and
Giulia Avanzato$^{2}$\thanks{Equal contribution}
\and
Soufiane Belharbi$^{1}$\footnotemark[1]
\and
Alexis Guichemerre$^{1}$\footnotemark[1]
\and
Luke McCaffrey$^{3}$
\and
Mohammadhadi Shateri$^{1}$
\and
Eric Granger$^{1}$
\and
$^{1}$ LIVIA, ILLS,  Dept. of Systems Engineering, ETS Montreal, Canada \\
$^{2}$ Dept. of Computer Engineering, University of Cagliari, Italy \\
$^{3}$ Goodman Cancer Research, Centre, Dept. of Oncology, McGill University, Canada
\and
$^{1}$\tt\small\{banafsheh.karimian.1, alexis.guichemerre.1, soufiane.belharbi, mohammadhadi.shateri, \\
\tt\small eric.granger\}@etsmtl.ca,
$^{2}${\tt\small g.avanzato@studenti.unica.it}, $^{3}$ luke.mccaffrey@mcgill.ca
}

\begin{document}
\maketitle
\setcounter{footnote}{0}
\input{sec/0_abstract}    
\input{sec/1_intro}

\input{sec/2_relatedwork}
\input{sec/3_method}
\input{sec/4_result}
\input{sec/5_conclusion}
{
    \small
    \bibliographystyle{ieeenat_fullname}
    \bibliography{main}
}
\clearpage
\appendix
\input{sec/apx_eval}
\input{sec/apx_dataset}
\input{sec/apx_models}
\input{sec/apx_algo}
\input{sec/apx_prompts}
\input{sec/apx_comp}
\end{document}

%% file: sec/0_abstract.tex
\begin{abstract}

Multimodal learning has shown promise in medical imaging, combining complementary modalities like images and text. Vision-language models (VLMs) capture rich diagnostic cues but often require large paired datasets and prompt- or text-based inference, limiting their practicality due to annotation cost, privacy, and compute demands. Crucially, available free unpaired external text, like pathology reports, can still provide complementary diagnostic cues if semantically relevant content is retrievable per image. To address this, we introduce \clipit, a novel framework that relies on rich unpaired text reports. Specifically, \clipit uses a CLIP model pre-trained on histology image–text pairs from a separate dataset to retrieve the most relevant unpaired textual report for each image in the downstream unimodal dataset. These reports, sourced from the same disease domain and tissue type, form pseudo-pairs that reflect shared clinical semantics rather than exact alignment. Knowledge from these texts is distilled into the vision model during training, while LoRA-based adaptation mitigates the semantic gap between unaligned modalities. At inference, only the vision model is used, keeping overhead low while still benefiting from multimodal training without requiring paired data in the downstream dataset. Experiments\footnote{\href{https://github.com/BanafshehKarimian/ModalityPairing/tree/main}{Github Link}} show that \clipit consistently improves classification accuracy over both unimodal and multimodal CLIP-based baselines in most cases, without the burden of per-dataset paired annotation or inference-time complexity.

\end{abstract}

%% file: sec/1_intro.tex
\section{Introduction}
\label{sec:intro}

Cancer diagnosis is among the most critical and challenging tasks \cite{kumaraswamy2022key}. To ensure a comprehensive understanding of cancer, medical professionals rely on the results of multiple modalities, such as histology images, genomic profiles, and pathology reports. Although machine and deep learning (ML/DL) models have shown great potential in advancing cancer analysis, unimodal models have limited ability to capture the complex, multifaceted nature of cancer, reducing model accuracy, robustness, and generalizability \cite{abdullakutty2024histopathology,sfdawsol,pixelcam}. Multimodal learning addresses these limitations by using the complementary strengths of each modality to improve performance \cite{ramachandram2017deep}. For instance, histology images have fine-grained morphological details, while pathology reports capture high-level clinical context. Combining these complementary modalities improves predictive accuracy, interpretability, and robustness, making models more resilient to noise and missing information \cite{8269806}.

Recent works \cite{abdullakutty2024histopathology,Lon_MuGI_MICCAI2024,qu2024multi} have shown that multimodal approaches have great potential for cancer detection and diagnosis, significantly pushing the boundaries of what ML can achieve in this field. A prominent class of recent multimodal models is vision-language models (VLMs), which jointly learn from paired image–text data to capture spatial and semantic information. These models, such as CLIP \cite{radford2021learning} and its medical variants, such as CONCH \cite{lu2024avisionlanguage}, which are trained on paired image–text datasets to jointly capture spatial and semantic information. Although their inference is often unimodal or prompt-based, these models depend on large-scale datasets with image-text pairs during training. However, in histopathology, obtaining such pairs for each dataset is expensive and time-consuming and requires strict privacy compliance and institutional approvals. Most publicly available datasets, such as PCAM \cite{Veeling2018-qh,veeling2018rotation}, BACH \cite{aresta2019bach}, and CRC \cite{Oliveira2021}, contain only histology images, lacking the paired textual annotations needed for such training, which limits the scalability of these approaches. 

In contrast, prompt-based methods \cite{Ngu_Towards_MICCAI2024} use short texts, handcrafted or template prompts at both training and inference time, bypassing the need for real reports altogether. While this approach reduces data collection costs, it is inherently limited in expressiveness and fails to reflect the nuanced, case-specific insights present in authentic pathology reports. Furthermore, these models are shown to be highly sensitive to minor changes in phrasing for pathology \cite{10943813}, leading to instability in clinical applications. Generative language models offer an alternative by synthesizing full reports given an image, but they are often proprietary, and expensive to run. Thus, both approaches face critical barriers in terms of scalability, reliability, and clinical applicability.

\begin{figure*}
\center
\includegraphics[width=0.99\textwidth]{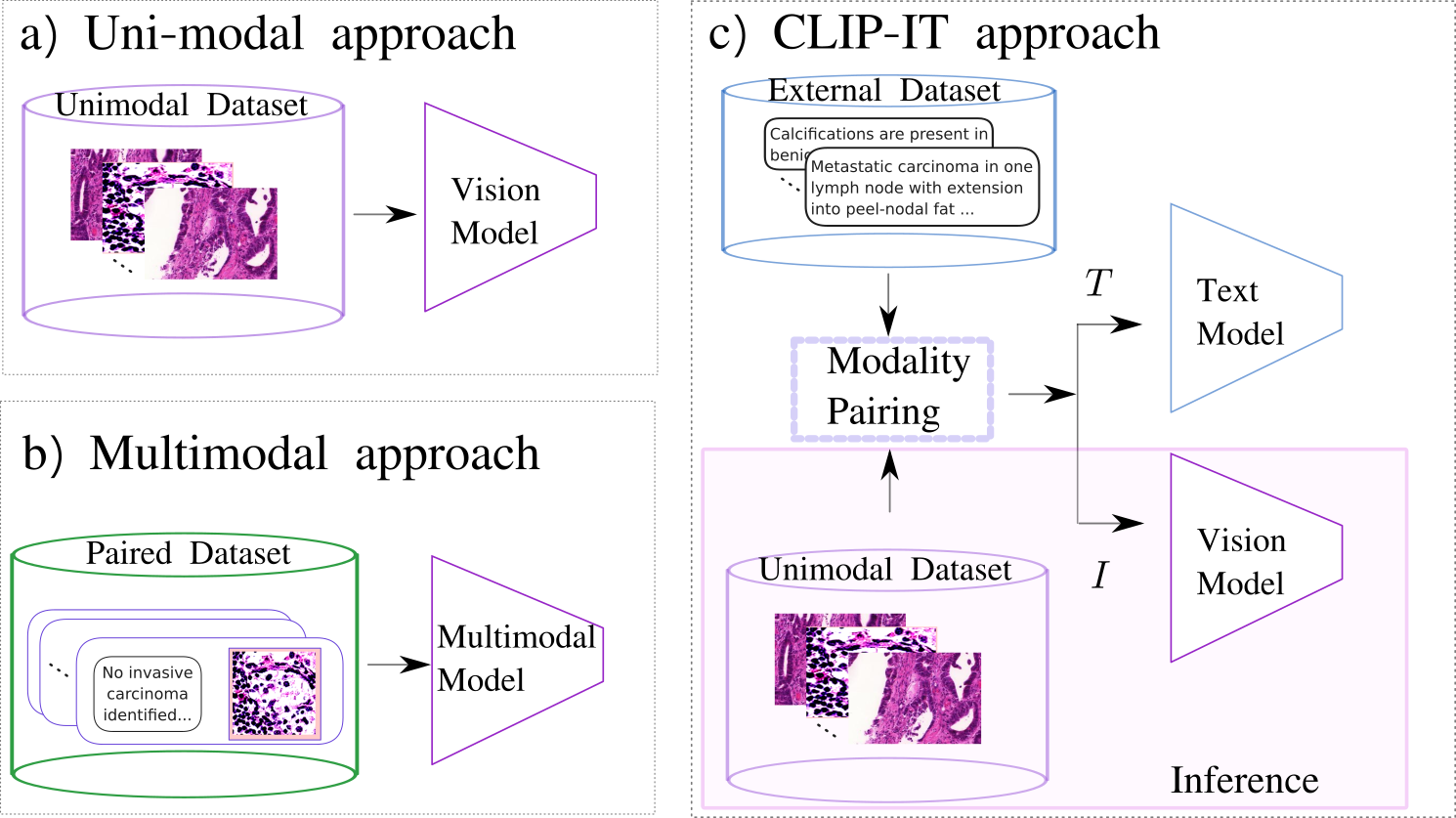}
\caption{Overview of four learning approaches: (a) unimodal setting (image-only), (b) paired multimodal setting, requiring aligned image–text pairs at both training and inference, (c) Prompt-based CLIP-style VLMs (e.g., CONCH), trained on paired data and needing text prompts at inference, and (d) the proposed \clipit setting that uses unpaired external reports for multimodal supervision during training, but supports lightweight unimodal inference of downstream dataset.} \label{fig:difference}
\end{figure*}

To address these challenges, we propose \clipit a method that enables multimodal training for unimodal image classification. \clipit uses freely available resources, i.e., unpaired textual reports from other datasets together with existing pretrained models, thus avoiding the need for paired annotations in each downstream dataset. This is especially beneficial in domains like pathology, where aligned datasets are scarce. The core idea is to consider unpaired but semantically relevant textual information as a form of privileged supervision, only available during training. Using a CLIP-based retrieval model, it automatically pairs each histology image with the most relevant textual report from another dataset of the same disease domain and tissue type. This forms pseudo-pairs that capture shared semantics rather than exact correspondence, enabling us to enrich unimodal datasets with complementary high-level information from text. Knowledge distillation is then used to transfer information from the text modality into the vision model, allowing efficient unimodal inference without requiring the text or text models at test time. \autoref{fig:difference} illustrates how \clipit setting compares to other paradigms: unlike traditional unimodal or fully paired multimodal models, and unlike CLIP-based approaches that rely on paired data during training and prompts at inference, \clipit performs multimodal training with unpaired data and enables efficient unimodal deployment without requiring paired annotations in the downstream dataset.

This work studies how multimodal learning can be used even when only unimodal datasets are available, an important direction for data-scarce domains like pathology, where collecting aligned multimodal samples is costly, time-consuming, and restricted by privacy regulations. By relying on semantically relevant external data as privileged information during training, our method enables the integration of diverse and heterogeneous data sources into medical AI pipelines. The contributions of this paper are as follows: \\
    \textbf{(1)} We introduce \clipit to improve histology classification, through multimodal learning, without the need for manually curated paired annotations of all the downstream datasets. It extends unimodal histology datasets by pairing them with histology reports from external datasets, creating a pseudo-paired multimodal dataset that includes potentially complementary textual information. \\
    \textbf{(2)} \clipit uses complementary text from another dataset as privileged information, using knowledge distillation to transfer information from the text modality to the vision model during training. This enables the use of multimodal information without requiring text or text-related models at test time, allowing for efficient unimodal inference. A LoRA-based adaptation is used to handle noise and misalignment in the pseudo-paired training signals.\\
    \textbf{(3)} Our extensive experiments show the effectiveness of \clipit in improving histology image classification of unimodal datasets. On the unimodal front, \clipit improves accuracy by up to $4.4\%$, $3.6\%$, and $1.5\%$ on PCAM, BACH, and CRC, respectively, all with minimal additional inference overhead. Notably, \clipit is a flexible framework that can be integrated with a variety of unimodal vision architectures, without requiring architectural modifications. This makes it broadly adaptable across standard vision backbones commonly used in histopathology. Furthermore, compared to state-of-the-art multimodal models specific to histology, such as CONCH \cite{lu2024avisionlanguage} and QUILTNet \cite{ikezogwo2023quilt}, \clipit achieves a higher accuracy across most cases, while maintaining comparable performance in the rest, and without requiring paired annotations for the downstream dataset or dual-modality inference.

%% file: sec/2_relatedwork.tex
\section{Related Work}
\label{sec:rw}
We summarize key trends and limitations of multimodal approaches for histopathology in following subsections.

\noindent\textbf{(a) Integration of Histology Images with Other Modalities:}
One commonly used modality alongside histology images is genomic data \cite{Lon_MuGI_MICCAI2024,Zha_Knowledgedriven_MICCAI2024,qu2024multi,Xio_MoME_MICCAI2024,Cai_Survival_MICCAI2024,Mej_Enhancing_MICCAI2024,yang2024spatial,Zha_DSCENet_MICCAI2024}. However, it is costly, often unavailable, and difficult to integrate due to high dimensionality and preprocessing demands, with studies using WSIs and genomic profiles \cite{Zha_Knowledgedriven_MICCAI2024,Cai_Survival_MICCAI2024} facing major data and compute challenges. Another increasingly explored modality is text, in the form of clinical reports or structured annotations derived from medical records \cite{Kim_LLMguided_MICCAI2024}. Recent works align histology images with such textual information to improve performance, especially for survival analysis with missing modalities \cite{qu2024multi}. For example, PathM3 \cite{Zho_PathM3_MICCAI2024} uses limited paired WSI–caption data to enhance MIL-based classification. Building on these efforts, contrastive VLMs have emerged as powerful frameworks that align histology images and text through large-scale pretraining. Prominent examples such as CONCH~\cite{lu2024avisionlanguage} and QUILTNet~\cite{ikezogwo2023quilt} are trained on extensive paired datasets and demonstrate strong performance across a wide range of pathology tasks, including classification, segmentation, and retrieval. However, their reliance on curated image–text pairs limits training scalability, and their zero-shot prompt-based performance remains suboptimal, motivating using unpaired data when per-dataset pairing is unavailable.


\noindent\textbf{(b)  Histology Image to Text Translation:} Recent work has explored generating textual descriptions from histology images, either to summarize WSIs \cite{Guo_HistGen_MICCAI2024,Che_WsiCaption_MICCAI2024} or extract structured visual features aligned with pathology hierarchies \cite{Wat_Hierarchical_MICCAI2024}. These methods require paired image–text datasets, which are costly to produce due to the need for expert annotation and validation. Moreover, training such generative models is computationally intensive, especially with large vision-language architectures \cite{huang2025survey}.

\noindent\textbf{(c)  Prompt-based Vision‑Language Models:} Among VLMs, short description or prompt‑based methods stand out for their simplicity and effectiveness. They have been applied to tasks like WSI-based genetic biomarker prediction \cite{zhang2024prompting}, data augmentation \cite{oh2024controllable}, adaptation of foundation models to pathology with task-specific visual and textual prompts \cite{Lu_PathoTune_MICCAI2024}, and detection of rare or novel diseases using disease-informed prompts and prototype learning~\cite{10723745}. A related method \cite{Ngu_Towards_MICCAI2024} was proposed that generates keyword-based short sentences, selecting top‑k prompts to derive more interpretable embeddings, which improves model explainability. However, prompt-based models rely heavily on handcrafted templates or keyword phrases, which may fail to capture the nuanced language and reasoning found in real clinical reports. Moreover, they are highly sensitive to prompt phrasing and offer limited coverage over complex diagnostic scenarios. Prior work~\cite{10943813} also highlights prompt sensitivity to phrasing, showing that minor variations can significantly impact model performance, especially in pathology. Thus, prompt-based tuning may fall short when trying to encode the full range of pathological descriptions necessary for robust diagnosis.

In summary, current multimodal approaches in histopathology often face significant limitations: text-generation models require paired annotations and introduce computational costs, and prompt-based VLMs are constrained by prompt sensitivity. Large-scale VLMs such as CONCH have demonstrated strong performance across multiple tasks, but rely on extensive paired training data and impose significant inference overhead due to large text encoders. These limitations motivate the need for \clipit, a lightweight alternative that still uses rich textual knowledge, via unpaired rich clinical data, but without requiring annotations for each downstream dataset or the high cost of a language backbone at inference time.

%% file: sec/3_method.tex
\section{Proposed \clipit Method}
\label{sec:poroposal}

\begin{figure*}
\centering
\includegraphics[width=0.8\textwidth]{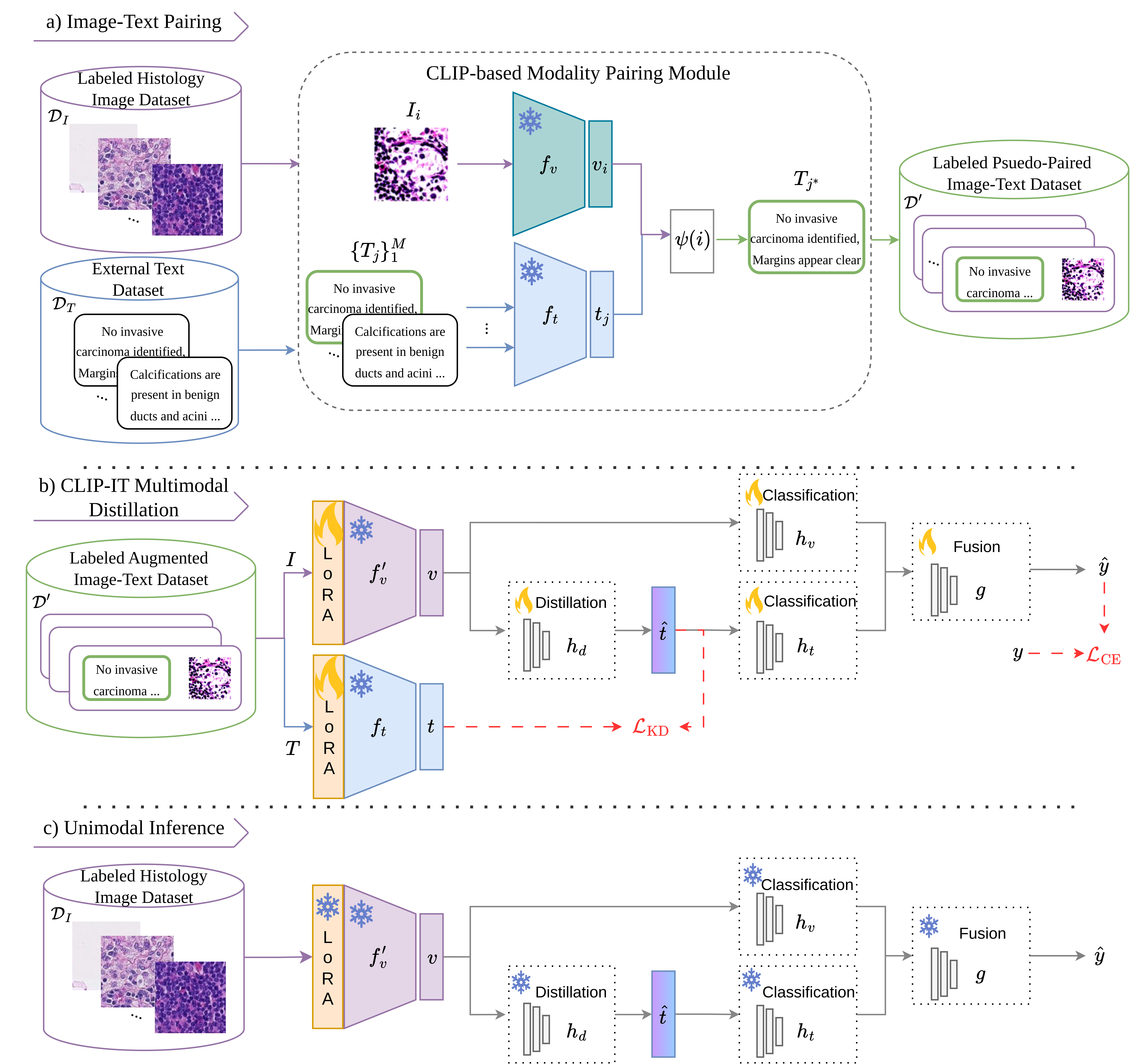}
\caption{Illustration of our \clipit method: a) Image-Text Modality Pairing: Each histology image is paired with the most semantically similar text report from an external unpaired dataset using a pretrained CLIP-based model, b) \clipit Multimodal Distillation: A joint model is trained using both vision and text encoders with a logit fusion mechanism and feature-level distillation. c) Unimodal Inference: only the vision encoder and the learned projection modules are used, enabling a lightweight and unimodal prediction pipeline.} 
\label{fig:proposal}
\end{figure*}

\clipit is proposed to enhance unimodal histology image classification using an external unpaired text modality, containing reports from the same domain, as privileged information during training. Although these reports are not aligned with the downstream dataset, they carry complementary domain knowledge, such as diagnostic terminology, patterns of disease progression, and clinical reasoning, that help the model learn richer semantic representations and improve generalization. \clipit retrieves the most semantically relevant external report for each image using an off-the-shelf CLIP-based model trained elsewhere on paired data. Importantly, this requires no paired annotations for the downstream dataset, since the pairing is performed automatically at retrieval time. Knowledge from these pseudo-pairs is distilled into the vision model, enabling it to learn from text during training while requiring only images at inference time. \autoref{fig:proposal} shows an overview of the \clipit pipeline, including image–text modality pairing, multimodal distillation, and unimodal inference. The remainder of this section details \clipit component. 

\subsection{Image-Text Modality Pairing}
Let us consider a unimodal histology dataset ${\mathcal{D}_I = \{(I_i, y_i)\}_{i=1}^{N}}$, composed of $N$ images and the corresponding image class label $y_i$ for each image,  $I_i$. 
We consider an external text dataset, ${\mathcal{D}_T = \{T_j\}_{j=1}^{M}}$, composed of $M$ histology medical reports, providing diagnostic observations and other clinical insights. 
The only assumption that we have for the two datasets, ${\mathcal{D}_T}$ and ${\mathcal{D}_I}$, is that they are relevant histology data, meaning they originate from the same organ type or disease domain (e.g., breast, colorectal), without the need to be explicitly paired at the sample level. Using the external dataset ${\mathcal{D}_T}$, we \emph{augment} our image dataset $\mathcal{D}_I$ from unimodal into a new multimodal dataset ${\mathcal{D}^{\prime}}$, containing vision-text-label pairs. To this end, an off-the-shelf CLIP model is used, pre-trained on paired histology images and text, composed of a vision encoder, ${f_v}$, and a text encoder, ${f_t}$. While these pretrained models depend on paired data at their origin, our contribution lies in showing that once trained, they can be reused as one-time resources to enable pseudo-pairing across many downstream unimodal datasets without further paired annotations. We define vision embeddings as $\mathbf{v}_i = f_v(I_i)$ for ${I_i \in \mathcal{D}_I}$ over images, and text embeddings as: $\mathbf{t}_j = f_t(T_j)$ for ${T_j \in \mathcal{D}_T}$ over the external text dataset. 

Given a vision embedding ${\mathbf{v}_i}$ of an image ${I_i}$ from ${\mathcal{D}_I}$, we use the CLIP-model capacity of pairing image-text to find the most relevant text ${T_{j^*}}$, with $j^*$ defined as follows, from the external dataset ${\mathcal{D}_T}$ to be paired with $I_i$. This is achieved by finding the text with the highest cosine similarity with the image in their feature space using $\psi$:
\begin{equation}
    \label{eq:1}
    j^* = \psi(i) = \arg\max_{j \in \{1, \ldots, M\}} \frac{\mathbf{v}_i \cdot \mathbf{t}_j}{\|\mathbf{v}_i\| \|\mathbf{t}_j\|}.
\end{equation}
A pseudo-paired dataset is constructed with the paired vision-text modality noted as ${\mathcal{D}^{\prime} = \{(I_i, T_{\psi(i)}, y_i)\}_{i=1}^{N}}$. This pairing procedure is illustrated in \autoref{fig:proposal} (part a). 
 


\subsection{CLIP-IT Multimodal Distillation}
A key challenge in leveraging unpaired text is how to transfer its information into a vision model, trained on an unpaired image-only dataset. To address this, we fine-tune the text encoder $f_t$ with a classification head $h_t$ on top, using the paired text–label data ${{(T_{\psi(i)}, y_i)}_{i=1}^{N}}$. This supervision guides the transformer's attention to focus on class-relevant regions within the text. The text classifier $h_t(\cdot)$ is trained to minimize the Cross-Entropy loss:
\begin{equation}
\mathcal{L}_{\text{CE}}(y_i, \hat{y_i}) = -\sum_{c=1}^C y_i^{(c)} \log \hat{y}_i^{(c)},
\end{equation}
where $\hat{y}_i = h_t(t_i)$, and $C$ is the total number of classes.

After fine-tuning the text encoder on paired reports,  
this textual supervision must be effectively transferred into the vision model. Since the text and image data originate from different datasets, they are not paired manually. Therefore, early fusion techniques, i.e., combine modalities at the input or feature level, are not the best option~\cite{ramachandram2017deep}. We instead adopt a late fusion technique that processes each modality separately and merges output predictions at the decision level. In particular, we employ a logit fusion module, $g$, that combines the class logits from the image model, $h_v(f'_v(I))$, and the text model, $h_t(f_t(T))$, requiring no modification to the vision backbones. This makes \clipit applicable to any vision model with any architecture.

While this multimodal training uses complementary information from both modalities to potentially enhance performance over unimodal baselines, the resulting model still depends on the text modality at inference time, introducing additional computational and architectural complexity. To bypass this, we propose to use knowledge distillation at the feature level between both modalities. In particular, we consider distilling the features of text modality into a branch of the vision model. This is achieved by training a module ${h_d(\mathbf{v}_i) = \hat{\mathbf{t}}_i}$ to predict text features $\hat{\mathbf{t}_i}$ from vision features $\mathbf{v}_i$ as shown in \autoref{fig:proposal} (part b). This module is trained by minimizing the following loss function: 
\begin{equation}
    \mathcal{L}_{\text{KD}}(\mathbf{t}_i, \hat{\mathbf{t}_i}) = 1 - \frac{\hat{\mathbf{t}_i} \cdot \mathbf{t}_i}{\|\hat{\mathbf{t}_i}\| \|\mathbf{t}_i\|}.    
\end{equation}

Overall, our model is trained using the following loss:
\begin{equation}
    \mathcal{L}(y_i, \hat{y}_i, t_i, \hat{t_i}) = \mathcal{L}_{\text{CE}}(y_i, \hat{y}_i) + \lambda \; \mathcal{L}_{\text{KD}}(t_i, \hat{t}_i),
    \label{eq:total}
\end{equation}
where ${\lambda}$ is a weighting coefficient. We denote the full multimodal model as $\mathcal{M}_{\theta_M} = \{f_t, f'_v, h_t, h_v, h_d, g\}$.

Given the lack of alignment between modalities, freezing pretrained encoders limits adaptation to the target task. However, full fine-tuning is expensive and sensitive to hyperparameters. To balance adaptability and efficiency, we adopt LoRA~\cite{hu2022lora}, which enables low-cost tuning by using lightweight trainable layers into the backbone.


\subsection{Unimodal Inference} 
After the multimodal distillation is complete, \clipit discards the need for the text encoder, $f_t$, during inference by relying only on the image-based components of the trained model, as shown in \autoref{fig:proposal} (part c). Specifically, the final model $\mathcal{M}_{\theta_U} = \{f'_v, h_d, h_t, h_v, g\}$ uses the distilled knowledge from the text modality, embedded in the auxiliary module $h_d$, which approximates text features from image embeddings. At test time, only an input image $I$ is passed through the vision encoder to obtain features $v = f'_v(I)$, which are transformed by $h_d$ into an estimated textual representation $\hat{t} = h_d(v)$. These representations are then fused at the logit level using the fusion module $g(h_t(\hat{t}), h_v(v))$ to generate the final prediction $\hat{y}$. This design ensures unimodal and efficient inference while benefiting from the rich semantics learned from text during training, without requiring text prompts or paired annotations at test time. Detailed algorithm of \clipit is provided in Supplementary Material.

%% file: sec/4_result.tex
\section{Results and Discussion}

\subsection{Experimental Methodology}  Methods were evaluated using three challenging histology image datasets: PCAM \cite{Veeling2018-qh}, BACH \cite{aresta2019bach}, and CRC \cite{Oliveira2021}. PCAM includes 327,680 breast lymph node patches ($96 \times 96$ at $10\times$ magnification, $0.97,\mu m/px$), labeled as tumor or normal. BACH contains 400 breast tissue patches ($2048 \times 1536$ at $20\times$, $0.42,\mu m/px$) labeled as normal, benign, in situ, or invasive carcinoma. CRC provides 107,180 colorectal tissue patches ($224 \times 224$ at $20\times$, $0.50,\mu m/px$) across 9 tissue types, including stroma and lymphocytes. For external text pairing, we use pathology reports from the TCGA dataset \cite{TCGA}, due to its wide tissue and cancer-type coverage and public accessibility. We filter TCGA reports by organ terms (e.g., ‘breast’, ‘colorectal’) to restrict domain overlap. For the modality pairing step, we use the CONCH model~\cite{lu2024avisionlanguage}. While CONCH was pretrained on paired histopathology data, we reuse it here as a one-time resource to retrieve semantically relevant reports for unpaired downstream datasets. Its domain-specific training and contrastive objectives empirically yield reliable pairings without additional manual annotations in the downstream datasets.
For the text model, the text encoder of CONCH is used, and for the vision model, the best backbones introduced in~\cite{gatopoulos2024eva} were used, i.e., the self-supervised pretrained DINO-L/14 and the supervised vanilla Vision Transformers (ViT-B/16, ViT-B/8, ViT-S/16, ViT-S/8), along with UNI~\cite{Chen2024}. The average accuracy is reported across three different vision backbone runs from \cite{gatopoulos2024eva} passed through our setting. Details of the models, datasets, evaluation metrics, used hardware, and hyperparameters are in the Supplementary Material.

\subsection{Comparison with State-of-the-Art Methods}

\autoref{tab:perf} presents the average classification accuracy and standard deviation (std) across several state-of-the-art vision backbones, trained either as a unimodal baseline, or with the proposed \clipit method. Evaluation is performed on three histopathology datasets: PCAM, BACH, and CRC. \clipit improves the unimodal baselines in most settings and maintains performance in a few other settings. For example, on PCAM, \clipit yields gains as high as $+4.4\%$ for ViT-B/8. On BACH, performance improvements are most notable for models like UNI and ViT-S/16, with gains up to $+2.9\%$. The variance in BACH stems from the base vision model's instability on this small dataset. Our method consistently improves or matches performance despite this variance, showing robustness to backbone fluctuations. Although CRC has strong unimodal baselines (e.g., ViT-B/16 with $95.9\%$), \clipit still offers modest improvements or preserves parity in most cases. No unimodal backbone experiences significant performance degradation when enhanced with \clipit. These results have a combined p-value of $2.65\times10^{-5}$, indicating a highly significant overall effect. 

\autoref{tab:perfmulti} compares \clipit with multimodal baselines such as CONCH and QUILTNet, both of which rely on paired image–text data and dual-modality inference. For a controlled evaluation, we isolate the vision encoder from each model and assess three variants: (i) a standalone classification head trained on image data only, (ii) a contrastively fine-tuned prompt-based multimodal model, and (iii) the same vision encoder enhanced by \clipit. Details of the setting, including the used prompts, are in the Supplementary Material. Despite using only unpaired data and supporting unimodal inference, \clipit surpasses contrastive multimodal approaches in most configurations, despite using unpaired text during training and unimodal inference. For instance, on BACH, \clipit outperforms the full contrastive version of CONCH by a large margin ($85.05\%$ vs. $60.78\%$) and shows better performance than QUILTNet and CONCH in both PCAM and BACH. These results illustrate that unpaired textual supervision, when leveraged with \clipit, can be effective. 

Interestingly, the performance gains vary by dataset. CRC, which has fewer relevant reports (only 376 colon reports vs. over 1000 breast reports), exhibits smaller improvements. Cosine similarity histograms between images and their retrieved reports are in the Supplementary Material, further supporting this observation. This highlights one limitation of our approach: performance is sensitive to the relevance and richness of the external textual data, even when unpaired. Overall, these results show that \clipit is robust across datasets and architectures, and it enables efficient integration of unpaired textual supervision without increasing inference-time complexity, given semantically rich and clinically meaningful text. 

\begin{table*}
\centering
\caption{Classification accuracy ($\pm$ std) averaged over three runs for unimodal vision backbones and their \clipit-enhanced counterparts across PCAM, BACH, and CRC datasets. Backbone naming follows \cite{gatopoulos2024eva} (DINO: self-supervised ViT-L/14, and ViT-B/S: supervised ViT with patch sizes 16 or 8).
The $\Delta$ columns indicate the relative accuracy gain of \clipit over the unimodal baseline.}
\label{tab:perf}
\renewcommand{\arraystretch}{1.3}
\resizebox{.95\linewidth}{!}{%
\begin{tabular}{l|ccc|ccc|ccc}
\toprule
\multirow{2}{*}{\shortstack{\textbf{Unimodal} \\ \textbf{Backbone}}} 
& \multicolumn{3}{c|}{\textbf{PCAM}} 
& \multicolumn{3}{c|}{\textbf{BACH}} 
& \multicolumn{3}{c}{\textbf{CRC}} \\
& Unimodal & \clipit & $\Delta$ 
& Unimodal & \clipit & $\Delta$
& Unimodal & \clipit & $\Delta$ \\
\midrule
UNI \cite{Chen2024}          
& $94.24 \pm 0.14$ & $\mathbf{95.49 \pm 0.27}$ & $\mathbf{+1.3}$ 
& $78.89 \pm 1.56$ & $\mathbf{81.79 \pm 1.98}$ & $\mathbf{+2.9}$ 
& $94.66 \pm 0.41$ & $\mathbf{95.92 \pm 0.07}$ & $\mathbf{+1.3}$ \\
DINO \cite{gatopoulos2024eva}     
& $88.88 \pm 0.75$ & $\mathbf{92.32 \pm 0.84}$ & $\mathbf{+3.4}$ 
& $84.26 \pm 2.30$ & $\mathbf{86.11 \pm 3.21}$ & $\mathbf{+1.8}$ 
& $94.40 \pm 0.40$ & $\mathbf{95.91 \pm 0.06}$ & $\mathbf{+1.6}$ \\
VITB-16 \cite{gatopoulos2024eva} 
& $88.13 \pm 1.09$ & $\mathbf{91.42 \pm 1.28}$ & $\mathbf{+3.3}$ 
& $80.78 \pm 2.14$ & $\mathbf{82.94 \pm 4.52}$ & $\mathbf{+2.1}$ 
& $\mathbf{95.86 \pm 0.25}$ & $95.67 \pm 0.18$ & $-0.2$ \\
VITS-16 \cite{gatopoulos2024eva} 
& $88.43 \pm 0.26$ & $\mathbf{90.93 \pm 1.12}$ & $\mathbf{+2.5}$ 
& $82.90 \pm 5.56$ & $\mathbf{84.64 \pm 6.68}$ & $\mathbf{+1.7}$ 
& $93.77 \pm 0.29$ & $\mathbf{95.34 \pm 0.20}$ & $\mathbf{+1.7}$ \\
VITB-8 \cite{gatopoulos2024eva}  
& $87.54 \pm 0.71$ & $\mathbf{91.92 \pm 0.87}$ & $\mathbf{+4.4}$ 
& $86.90 \pm 2.32$ & $\mathbf{87.06 \pm 1.22}$ & $\mathbf{+0.2}$ 
& $\mathbf{95.71 \pm 0.11}$ & $95.66 \pm 0.53$ & $-0.1$ \\
VITS-8 \cite{gatopoulos2024eva}  
& $87.90 \pm 0.61$ & $\mathbf{90.24 \pm 0.90}$ & $\mathbf{+2.3}$ 
& $81.01 \pm 3.04$ & $\mathbf{82.55 \pm 1.57}$ & $\mathbf{+1.5}$ 
& $\mathbf{95.03 \pm 0.13}$ & $94.80 \pm 0.60$ & $-0.2$ \\
\bottomrule
\end{tabular}
}
\end{table*}
\begin{table*}
\centering
\caption{Comparison of \clipit with other multimodal backbones (CONCH, QUILTNet) and only their vision models.}
\label{tab:perfmulti}
\renewcommand{\arraystretch}{1.3}
\resizebox{.95\linewidth}{!}{%
\begin{tabular}{l|ccc|ccc|ccc}
\toprule
\multirow{2}{*}{\shortstack{\textbf{Multimodal} \\ \textbf{Backbone}}} 
& \multicolumn{3}{c|}{\textbf{PCAM}} 
& \multicolumn{3}{c|}{\textbf{BACH}} 
& \multicolumn{3}{c}{\textbf{CRC}} \\
& \clipit & Contrastive & Vision 
& \clipit & Contrastive & Vision 
& \clipit & Contrastive & Vision \\
\midrule
CONCH \cite{lu2024avisionlanguage} 
& $\mathbf{93.61 \pm 0.44}$ & $92.67 \pm 1.40$ & $91.75 \pm 2.57$ 
& $\mathbf{85.05 \pm 0.64}$ & $60.78 \pm 0.29$ & $67.25 \pm 4.34$ 
& $94.89 \pm 0.61$ & $\mathbf{95.58 \pm 0.40}$ & $95.12 \pm 0.47$ \\
QUILTNet \cite{ikezogwo2023quilt} 
& $\mathbf{91.83 \pm 2.37}$ & $89.82 \pm 0.62$ & $90.44 \pm 0.65$ 
& $\mathbf{65.50 \pm 1.84}$ & $55.82 \pm 5.86$ & $63.81 \pm 2.16$ 
& $94.83 \pm 0.94$ & $\mathbf{95.37 \pm 0.17}$ & $94.60 \pm 0.65$ \\
\bottomrule
\end{tabular}
}
\end{table*}

\subsection{Performance vs. Efficiency Trade-Off}

\begin{figure*}
    \centering
    {\includegraphics[width=0.95\textwidth]{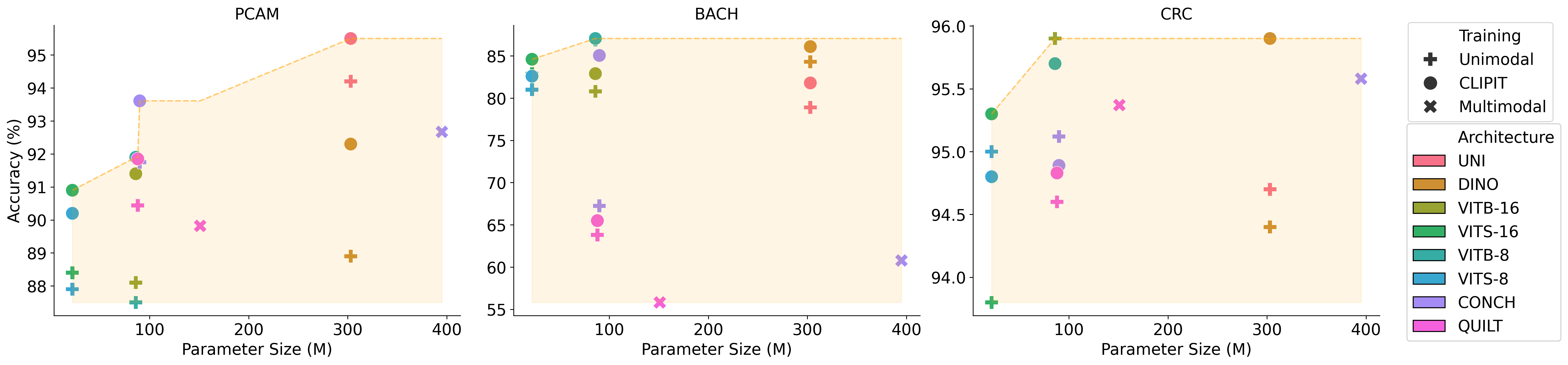}}
\caption{Pareto frontier plots showing the trade-off between model accuracy and parameter size across three histology datasets. Each point style represents a model configuration (Unimodal, \clipit, or Multimodal), with color indicating the architecture. \clipit consistently pushes unimodal models closer to or onto the frontier, offering an efficient alternative to heavier multimodal baselines.
} 
\label{fig:pareto}
\end{figure*}

To understand the trade-off between model complexity and accuracy, we analyze the parameter size versus performance across all models and configurations. \autoref{fig:pareto} presents Pareto frontier plots for PCAM, BACH, and CRC, comparing the three setting: \ding{108}) the vision encoder enhanced with \clipit, \ding{58}) the same encoder with a standalone classification head trained on image data, and \ding{54}) a contrastively fine-tuned multimodal model. Across datasets, \clipit models lie on or near the Pareto frontier, indicating that it achieves the best trade-off between model size and classification accuracy compared to other models of similar complexity. In particular, \clipit outperforms larger unimodal baselines and approaches and mainly surpasses the performance of heavier multimodal models, all while maintaining a significantly smaller parameter size.


\subsection{Complementarity of Text Supervision}
\begin{figure}
\centering
\includegraphics[width=0.50\textwidth, clip, trim=0.4cm 0.3cm 0.1cm 0.0cm]{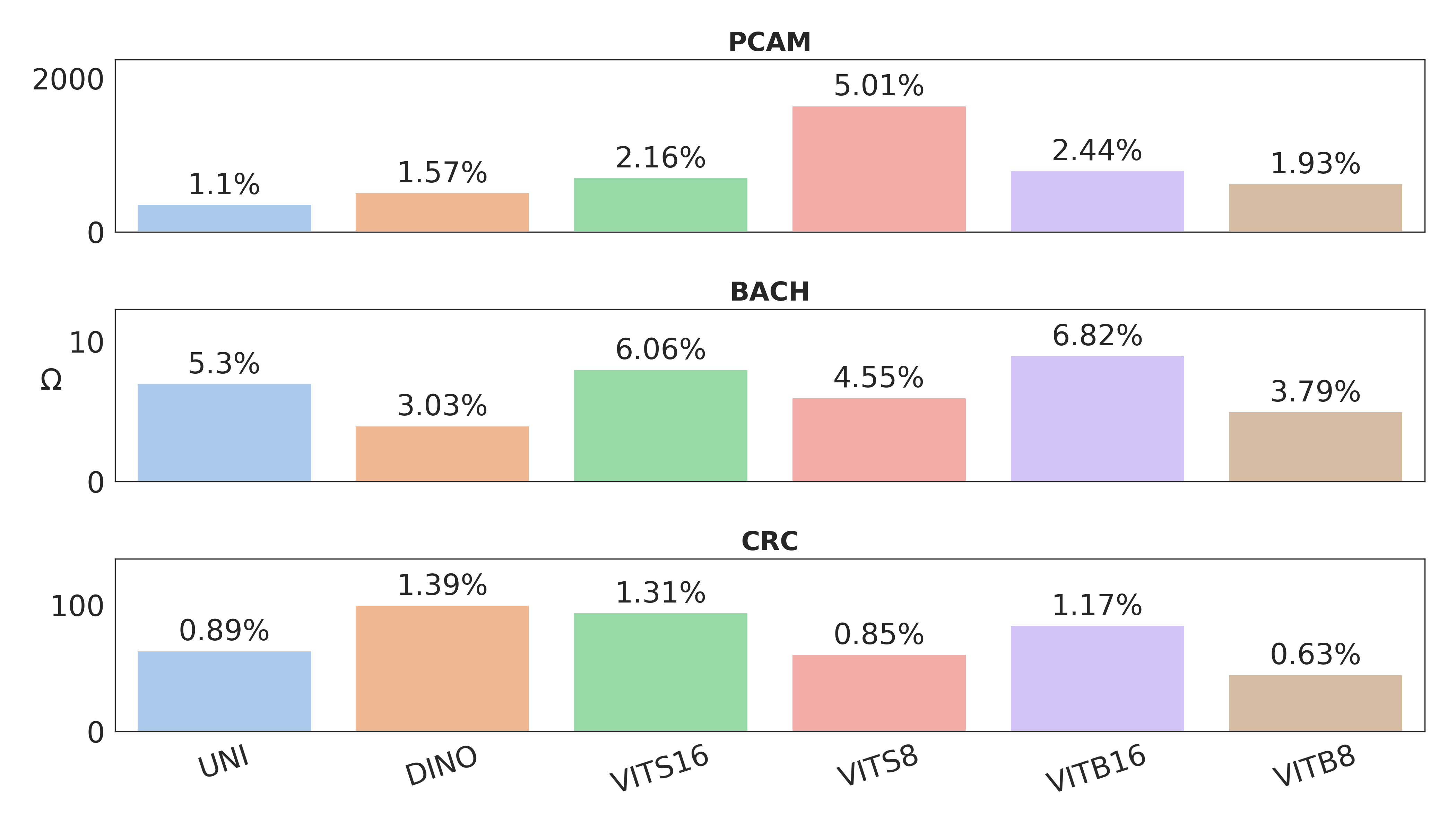}
\caption{Histogram of $\mathrm{\Omega}$ scores (\autoref{eq:omega}) across datasets and backbones, showing the $\%$ of samples correctly classified by text but missed by vision, i.e., the complementary info of text. Bars denote models, with numeric values $(\mathrm{\Omega}\times100)$ above each. Higher scores indicate greater potential benefit from textual supervision.
} 
\label{fig:gain}
\end{figure}

\clipit results were further analyzed by inspecting the source of performance gain in \autoref{tab:perf}. To this end, the following ${\mathrm{\Omega}}$ measure is considered to capture the impact of text modality in terms of classification accuracy, defined as: 
\begin{equation}
     \mathrm{\Omega}(Y_V, Y_T) = \sum_{i=1}^{N} \mathbbm{1}_{\{y_i^t = y_i \bigwedge y_i^v \not= y_i\}},
    \label{eq:omega}
\end{equation}
where $Y_V = \{y^v_i\}_{i=1}^N$, $Y_T = \{y^t_i\}_{i=1}^N$, with $y_i$, $y_i^v$, and $y_i^t$ are the $i_{th}$ true label, prediction of the vision model, and that of the text model, while $\mathbbm{1}_{\{\cdot\}}$ is the indicator function, which is equal to one if the condition inside is true, and zero otherwise. This measure captures the complementary, class-discriminative signals that the text modality contributes. It counts the number of samples the text model can correctly classify, while the vision model fails. \autoref{fig:gain} shows the value of $\mathrm{\Omega}$ for the different datasets and backbones. We can see that for all the models, there is a notable number of cases where the text modality can provide the correct class to the vision model. However, we observe a performance gap across backbones and datasets, with \clipit providing the most benefit on PCAM. This dataset contains a large number of training samples and exhibits high intra-class variability, which increases the difficulty of distinguishing between certain classes using visual features alone. In such cases, visual features alone may be ambiguous, and the external text reports provide complementary clinical cues that help disambiguate similar classes. This is supported by higher $\mathrm{\Omega}$ values, indicating a stronger contribution from the text modality on PCAM. These results suggest that \clipit is especially beneficial in settings where visual features are ambiguous and textual cues can serve as discriminative signals.

\subsection{Ablation Studies}
\begin{figure}
    \centering
\includegraphics[width=0.48\textwidth, clip, trim=0.4cm 0.3cm 0.1cm 0.0cm]{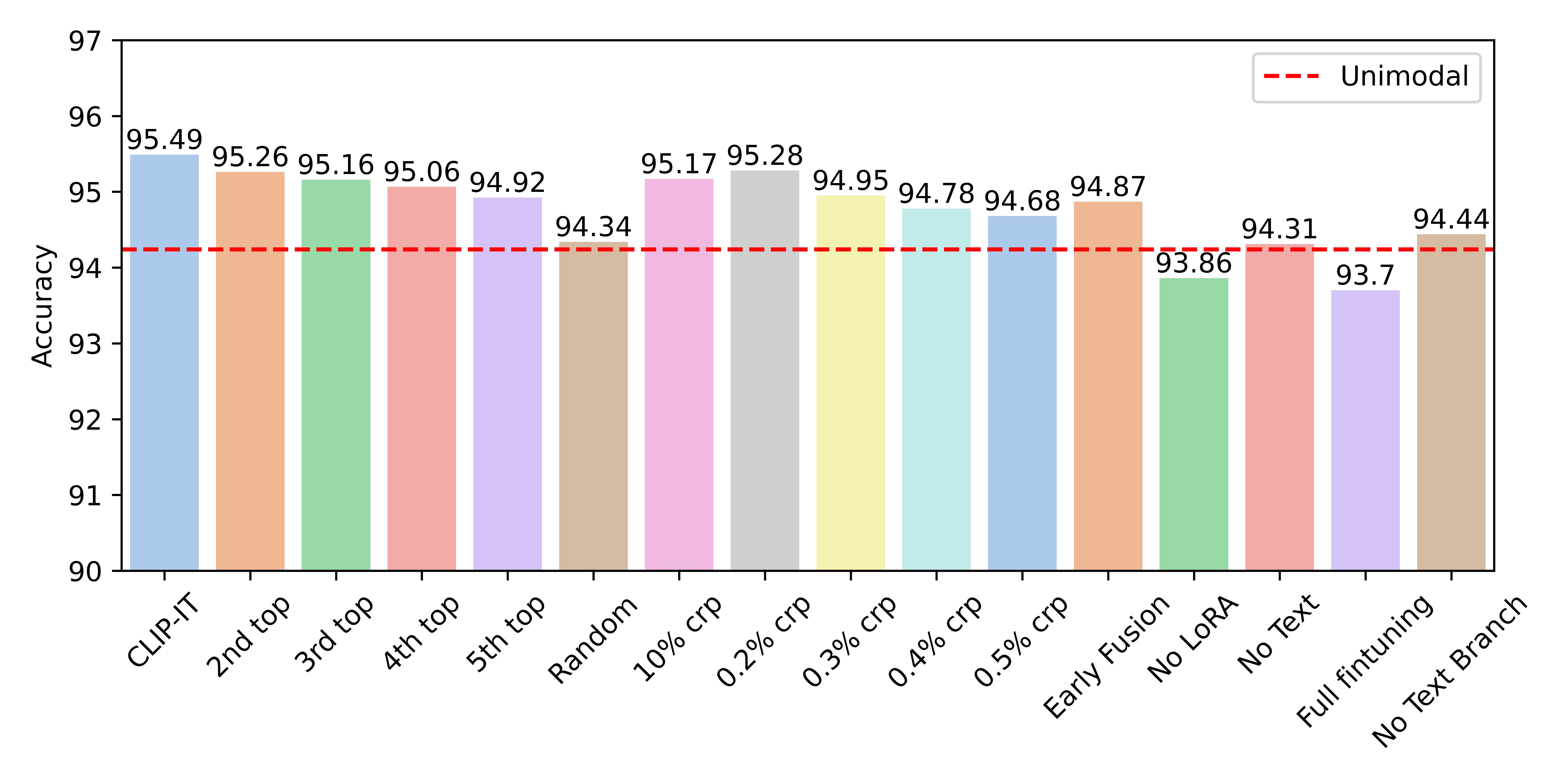} 
\caption{Ablation study results showing the classification accuracy of various configurations on UNI and PCAM. The bars represent modifications, including pairing strategies (2nd–5th top), text corruption by k\% word removal, early fusion, full fine-tuning, and component removals. The dashed line is the unimodal baseline.}
\label{fig:ablation}
\end{figure}

To understand each component in \clipit, we conduct a series of ablation studies summarized in \autoref{fig:ablation}.

\noindent - \textbf{Impact of LoRA:} First, we assess the impact of LoRA fine-tuning by removing it from the training setup. The performance drops from 95.49\% to 93.86\%. This gap highlights that LoRA is essential for adapting the model to noisy pseudo-pairs without disrupting pretrained representations. Unlike full fine-tuning, which can destabilize large vision and text encoders, LoRA applies low-rank updates to selected layers, allowing efficient adaptation while preserving the robustness of the original pretrained weights.

\noindent - \textbf{Comparing late vs. early fusion:} We compared the performance of our late fusion approach against an early fusion strategy, using a fully connected layer. As shown in \autoref{fig:ablation}, early fusion achieves 94.87\% accuracy on PCAM with UNI, whereas our late-fusion design, CLIP-IT, reaches 95.49\%. This difference results from early fusion being most suitable for aligned modalities \cite{ramachandram2017deep}, while in our setting the modalities are not perfectly aligned.

\noindent - \textbf{Text-to-vision distillation:} To assess the impact of having a dual branch, we evaluated a simplified configuration where the textual embedding is distilled directly into the vision embedding,  removing the additional text branch (see. \autoref{fig:ablation}). This setting achieves 94.44\% accuracy, slightly higher than the unimodal baseline (94.24\%) but still below our dual-branch framework (95.49\%). The gap can be explained by the fact that pseudo-pairs introduce noise into the image embeddings, which may confuse the model when text features are injected directly into the vision backbone. By maintaining separate branches, the vision encoder preserves the main discriminative features, while the text branch adds complementary cues when beneficial.

\noindent - \textbf{Impact of architectural modifications:}
We test whether improvements stem from architecture alone by training with the added modules but without text or distillation (see. \autoref{fig:ablation}). This setup yields 94.31\%, nearly identical to the vision backbone (94.24\%) yet well below full \clipit, showing that the gains are driven by textual supervision rather than structural changes.

\noindent - \textbf{Robustness to input noise:}
We randomly remove a fraction (from 0 to 50\%) of words from each report during training. Up to 30\% word dropout, the model maintains high accuracy (e.g., 94.95\% at 30\%), showing resilience to moderate textual degradation. Beyond this threshold, accuracy declines more noticeably, approaching the unimodal baseline, suggesting that while \clipit can extract value from partial supervision, it still requires a meaningful portion of the text to deliver improvements.

\noindent - \textbf{Impact of pairing quality and irrelevant text:} We also study the effect of pairing quality. In particular, we evaluate performance using the 2nd to 5th most similar reports (based on cosine similarity), as well as entirely random pairings. Accuracy gradually declines as we move from top-ranked to lower-ranked reports, confirming that better semantic alignment between images and text results in better supervision. Notably, when using random text pairings, the performance drops to unimodal (94.34\%), showing the importance of semantically relevant matches. These results show that the gains of \clipit stem from informative text and its lightweight distillation design \footnote{Further analyses (visualizations, generalization, retrieval, text supervision, and distillation) are provided in the Supplementary Material.}.

\subsection{Test-Time Computational Cost}
We compared \clipit’s efficiency to CONCH and its unimodal vision encoder. The unimodal baseline costs ~17 GFLOPs per forward pass, \clipit\ adds only 0.001 GFLOPs, whereas CONCH costs ~505 GFLOPs. On an RTX A6000 (batch=1), latency is 33.3 ms (unimodal), 43.1 ms for \clipit\ (+29\%), and 158.5 ms for CONCH (+376\%). Thus, \clipit\ delivers multimodal gains at near-unimodal compute, supporting real-time deployment.

%% file: sec/5_conclusion.tex
\section{Conclusion}

We introduced \clipit, a multimodal framework that improves histology image classification without requiring manually curated image–text pairs for each downstream dataset. By aligning each image with a semantically relevant external report using a pretrained CLIP-based retrieval model, \clipit forms pseudo-pairs and distills textual knowledge into the vision model. This enables effective multimodal supervision during training while allowing lightweight unimodal inference. Experiments show that \clipit can improve unimodal baselines and often improves fully multimodal models, all with negligible overhead at inference time. In this way, \clipit offers a cost-effective alternative for domains where collecting paired annotations is infeasible.

\noindent \textbf{Limitations and Future Work:} Despite its strong performance, \clipit has smaller gains for datasets with sparse or noisy reports (e.g., CRC). Mainly tested for classification, extending \clipit to tasks like segmentation is an important future direction. Also, using other modalities, such as genomic data, could further enhance multimodal computational pathology.

%% file: sec/apx_eval.tex
\section{Complementary Details}
\begin{itemize}
    \item \textbf{Evaluation Metrics:} Our primary evaluation metric is accuracy. For each dataset, we conduct three independent runs with different random seeds and report the mean accuracy along with the standard deviation to account for performance variability and ensure statistical robustness. Furthermore, to assess the statistical significance of the reported performance improvements across models and datasets, we use Fisher’s Combined Probability Test. In addition, we evaluate the computational cost in terms of floating point operations (FLOPs) that provide a hardware-agnostic estimate of inference efficiency for a fair comparison of computational complexity.
    \item \textbf{Hardware:} All training was performed on NVIDIA RTX A6000 and NVIDIA A100 GPUs.
    \item \textbf{Training Details:} The fixed hyperparameters we used are shown in ~\autoref{tab:training-details}. For the hyperparameter search, we focus on LoRA's hyperparameters, i.e., LoRA rank ($r \in \{8, 16, 32\}$), LoRA scale ($\alpha \in \{2, 4, 8\}$), and LoRA targets (a combination of Linear, 2D Convolutional, and Embedding layers), in addition to $\lambda$. 
    \begin{table}[h]
    \centering
    \caption{Training Hyperparameters}
    \begin{tabular}{ll}
    \toprule
    \textbf{Parameter} & \textbf{Value} \\
    \midrule
    Optimizer          & Adam \\
    Learning Rate      & $1 \times 10^{-3}$ \\
    Weight Decay       & $1 \times 10^{-4}$ \\
    Batch Size         & 64 \\
    Epochs (PCAM, CRC) & 1 \\
    Epochs (BACH)      & 25 \\
    \bottomrule
    \end{tabular}
    \label{tab:training-details}
    \end{table}

\end{itemize}

%% file: sec/apx_dataset.tex
\section{Datasets}
\label{Apx:data}
In this work, we evaluate our method on three publicly available histopathology image datasets: PCAM, BACH, and CRC. These datasets are chosen to represent a diverse set of tissue types, magnification levels, and classification difficulties. PCAM has the most number of patches, while BACH has a small number of patches , 400. For PCAM, we used the offical train-test-validation set and for BACH and CRC we use 10\% of the train set as validation. A summary of the dataset characteristics is provided in \autoref{tab:dataset-details}. All datasets are used exclusively for image classification, with no textual metadata. To enable modality pairing, external histopathology reports are sourced from the TCGA dataset, which provides rich textual descriptions for related tissue types.

\begin{table*}[h]
\centering
\caption{Summary of the histopathology image datasets used in this study. The datasets vary in organ domain, number of patches, image resolution, magnification, and number of classes.}
\begin{tabular}{|c|l|l|l|l|l|c|}
\hline
\textbf{Dataset} & \textbf{Domain} & \textbf{Patch \#} & \textbf{Patch Size} & \textbf{Magnification} & \textbf{Classes}  & \textbf{Class Names} \\
\hline
PCAM & Breast & $327,680$ & $96 \times 96$ & $10\times$ ($0.97\,\mu\text{m}/\text{px}$) & 2 & \makecell[l]{0: Normal \\ 1: Tumor}\\
\hline
BACH & Breast & $400$& $2048 \times 1536$ & $20\times$ ($0.42\,\mu\text{m}/\text{px}$) & 4 & \makecell[l]{1: Normal\\ 2: Benign \\ 3: In situ carcinoma \\ 4: Invasive carcinoma}\\
\hline
CRC  & Colorectal & $107,180$& $224 \times 224$ & $20\times$ ($0.50\,\mu\text{m}/\text{px}$) & 9 & \makecell[l]{0: Adipose tissue\\ 1: Background \\2: Debris (necrosis, mucus, hemorrhage) \\ 3: Lymphocytes \\ 4: Mucus \\ 5: Smooth muscle \\ 6: Normal colon mucosa\\ 
7: Cancer-associated stroma \\ 
8: Colorectal adenocarcinoma epithelium} \\
\hline
\end{tabular}
\label{tab:dataset-details}
\end{table*}

%% file: sec/apx_models.tex
\section{Models}
\label{Apx:model}
We evaluate our method using a variety of vision transformer models with diverse capacities, ranging from lightweight backbones, such as ViT-S/16, to large-scale multimodal frameworks, such as CONCH. \autoref{tab:model-summary} summarizes the models and their parameter sizes. For DINO ViT-L/14, ViT-S/16, ViT-S/8, ViT-B/16, and ViT-B/8 we use the pathology fine-tuned version \cite{gatopoulos2024eva}. Additionally, we employ state-of-the-art histology specialized multimodal vision-language models: CONCH~\cite{lu2024avisionlanguage} and QUILTNet~\cite{ikezogwo2023quilt}. While these are trained multimodally, we also extract and evaluate their vision backbones separately to isolate the contribution of vision-only learning from paired training. 
\begin{table}[h]
\centering
\caption{List of vision and multimodal models used in our experiments, along with their parameter sizes. We include both standard vision backbones (e.g., ViT-B/16, ViT-S/8) and large-scale multimodal models (e.g., CONCH, QUILTNet).}
\begin{tabular}{|l|c|}
\hline
\textbf{Model} & \textbf{\# Parameters} \\
\hline
CONCH \cite{lu2024avisionlanguage} & 395M \\
\hline
UNI \cite{Chen2024} & 303M \\
\hline
DINO ViT-L/14 \cite{gatopoulos2024eva} & 303M \\
\hline
QUILTNet \cite{ikezogwo2023quilt} & 151M \\
\hline
CONCH Vision Backbone \cite{lu2024avisionlanguage} & 90M \\
\hline
QUILTNet Vision Backbone \cite{ikezogwo2023quilt} & 88M \\
\hline
ViT-B/16 \cite{gatopoulos2024eva} & 86M \\
\hline
ViT-B/8 \cite{gatopoulos2024eva} & 86M \\
\hline
ViT-S/16 \cite{gatopoulos2024eva} & 22M \\
\hline
ViT-S/8 \cite{gatopoulos2024eva} & 22M \\
\hline
\end{tabular}
\label{tab:model-summary}
\end{table}

%% file: sec/apx_algo.tex
\section{Algorithms}

We outline the full training pipeline of \clipit in Algorithms~\ref{algo:pairing}–\ref{algo:clipit}. First, each histology image is pseudo-paired with the most semantically relevant external report using CLIP-based similarity (Algorithm~\ref{algo:pairing}). These pseudo-pairs are then used to train a multimodal model that learns to predict textual features from image representations and combine them for classification (Algorithm~\ref{algo:multimodtraining}). Finally, after training, the text encoder is discarded, and the resulting model retains only the vision branch and learned mappings, enabling efficient unimodal inference while benefiting from multimodal supervision (Algorithm~\ref{algo:clipit}).

\label{Apx:algo}
\begin{algorithm}[t]
\caption{Modality Pairing}
\label{algo:pairing}
\begin{algorithmic}[1]
\REQUIRE Image dataset $\mathcal{D}_I = \{(I_i, y_i)\}_{i=1}^N$; \\
    external text dataset $\mathcal{D}_T = \{T_j\}_{j=1}^M$
\ENSURE Paired dataset $\mathcal{D}' = \{(I_i, T_{\psi(i)}, y_i)\}_{i=1}^N$

\STATE Filter $\mathcal{D}_T$ with organ-specific keywords
\FOR{each $T_j \in \mathcal{D}_T$}
    \STATE $\mathbf{t}_j \leftarrow \frac{f_t(T_j)}{\|f_t(T_j)\|}$ \COMMENT{Normalized text embeddings}
\ENDFOR
\FOR{each $I_i \in \mathcal{D}_I$}
    \STATE $\mathbf{v}_i \leftarrow \frac{f_v(I_i)}{\|f_v(I_i)\|}$ \COMMENT{Normalized image embedding}
    \STATE $j^* \leftarrow \arg\max_{j} \frac{\mathbf{v}_i \cdot \mathbf{t}_j}{\|\mathbf{v}_i\| \|\mathbf{t}_j\|}$
    \STATE Add $(I_i, T_{j^*}, y_i)$ to $\mathcal{D}'$
\ENDFOR
\RETURN $\mathcal{D}'$
\end{algorithmic}
\end{algorithm}
\begin{algorithm}[t]
\caption{Multimodal Distillation}
\label{algo:multimodtraining}
\begin{algorithmic}[1]
\REQUIRE Paired dataset $\mathcal{D}' = \{(I_i, T_i, y_i)\}_{i=1}^N$; multimodal model: $\mathcal{M}_{\theta_M} = \{f_t, f'_v, h_t, h_v, h_d, g\}$ 
\ENSURE Trained multimodal model, $\mathcal{M}_{\theta_M}$

\FOR{each batch $(I^b, T^b, y^b)$ sampled from $\mathcal{D}'$ with \\$I^b = \{I_i^b\}_{i=1}^{N_B},\; T^b = \{T_i^b\}_{i=1}^{N_B},\; y^b = \{y_i^b\}_{i=1}^{N_B}$\\}
    \STATE $\mathbf{v}^b \leftarrow f'_v(I^b)$ \COMMENT{Image features}
    \STATE $\mathbf{t}^b \leftarrow f_t(T^b)$ \COMMENT{Text features}
    \STATE $\hat{\mathbf{t}}^b \leftarrow h_d(\mathbf{v}^b)$ \COMMENT{Predicted text from image}
    \STATE $\hat{y}^b \leftarrow g(h_t(\hat{\mathbf{t}}^b), h_v(\mathbf{v}^b))$ \COMMENT{Predicted label}
    \STATE Update $\theta_M$: 
    \\ \centerline{
    $\theta_M \leftarrow \theta_M - \eta \frac{1}{N_B} \sum_{i=1}^{N_B} \mathcal{L}\big(y_i^b, \hat{y}_i^b,\; t_i^b,\; \hat{t}_i^b\big)$
    }
\ENDFOR
\RETURN Trained $\mathcal{M}_{\theta_M}$
\end{algorithmic}
\end{algorithm}
\begin{algorithm}[t]
\caption{\clipit}
\label{algo:clipit}
\begin{algorithmic}[1]
\REQUIRE Image dataset $\mathcal{D}_I = \{(I_i, y_i)\}_{i=1}^N$; text dataset $\mathcal{D}_T = \{T_j\}_{j=1}^M$; a vision encoder $f'_v$; text encoder text-encoder $f_t$
\ENSURE Trained unimodal model: \\ \centerline{$\mathcal{M}_{\theta_U} = \{f'_v, h_t, h_v, h_d, g\}$}

\STATE $\mathcal{D}' \leftarrow$ \textbf{Modality Pairing}($\mathcal{D}_I$, $\mathcal{D}_T$)
\STATE Fine-tune text-encoder $f_t$ using text-label pairs $(T_i, y_i)$ from $\mathcal{D}'$
\STATE Initialize multimodal model: \\ \centerline{$\mathcal{M}_{\theta_M} = \{f_t, f'_v, h_t, h_v, h_d, g\}$} 
\FOR{each epoch}
    \STATE $\mathcal{M}_{\theta_M} \leftarrow$ \textbf{Multimodal Distillation}( $\mathcal{D}'$, $\mathcal{M}_{\theta_M}$)
\ENDFOR
\STATE Extract unimodal model: $\mathcal{M}_{\theta_U} \leftarrow \{f'_v, h_t, h_v, h_d, g\}$
\RETURN $\mathcal{M}_{\theta_U}$ for unimodal inference
\end{algorithmic}
\end{algorithm}

%% file: sec/apx_prompts.tex
\section{Prompts}
For the fine-tuning of multimodal models, specifically CONCH~\cite{lu2024avisionlanguage} and QUILTNet~\cite{ikezogwo2023quilt}, we adopt a contrastive learning strategy. We follow the procedure outlined in their original works, where image and text pairs are jointly optimized using a contrastive loss function to bring semantically aligned vision-text pairs closer in the shared embedding space, while pushing apart unrelated pairs. This objective encourages the models to learn rich cross-modal representations that align visual features with natural language descriptions. The text we used is the class name used with the prompts.

During inference, we evaluate the models using predefined textual prompts for each class. Given a set of candidate class labels, we replace "CLASSNAME" with each label in templates to form descriptive textual prompts. These prompts are encoded using the text encoder and compared to the image embeddings to compute similarity scores. The final prediction is made using template ensemble averaging, where the logits or similarities across templates are aggregated for each class.

We use the official prompt templates provided by the authors of each model:

\begin{itemize}
    \item \textbf{CONCH Prompts}: 22 natural language templates that reflect varied clinical and descriptive phrasings, such as:
    \begin{itemize}
        \item \textit{"an H\&E image of CLASSNAME."}
        \item \textit{"CLASSNAME."}
        \item \textit{"a photomicrograph showing CLASSNAME."}
        \item \textit{"a photomicrograph of CLASSNAME."}
        \item \textit{"an image of CLASSNAME."}
        \item \textit{"an image showing CLASSNAME."}
        \item \textit{"an example of CLASSNAME."}
        \item \textit{"CLASSNAME is shown."}
        \item \textit{"this is CLASSNAME."}
        \item \textit{"there is CLASSNAME."}
        \item \textit{"a histopathological image showing CLASSNAME."}
        \item \textit{"a histopathological image of CLASSNAME."}
        \item \textit{"a histopathological photograph of CLASSNAME."}
        \item \textit{"a histopathological photograph showing CLASSNAME."}
        \item \textit{"shows CLASSNAME."}
        \item \textit{"presence of CLASSNAME."}
        \item \textit{"CLASSNAME is present."}
        \item \textit{"an H\&E stained image of CLASSNAME."}
        \item \textit{"an H\&E stained image showing CLASSNAME."}
        \item \textit{"an H\&E image showing CLASSNAME."}
        \item \textit{"CLASSNAME, H\&E stain."}
        \item \textit{"CLASSNAME, H\&E."}
    \end{itemize}
    
    \item \textbf{QUILTNet Prompts}: 
    \begin{itemize}
        \item \textit{"a histopathology slide showing CLASSNAME"}
        \item \textit{"histopathology image of CLASSNAME"}
        \item \textit{"pathology tissue showing CLASSNAME"}
        \item \textit{"presence of CLASSNAME tissue on image"}
    \end{itemize}
\end{itemize}

The final prediction is made by selecting the class with the highest average similarity across its corresponding prompt variants.

%% file: sec/apx_comp.tex
\section{Complementary Results}
\subsection{Reports or prompts?}
To answer this question, we compare the performance of a classification layer on top of our text encoder, with different texts. On one side we have the reports we pair using \clipit, and on the other side, we have the prompts used for using our text-encoder (Conch text-encoder) in a CLIP-based approach. To do so, instead of reports, we pair the images of each dataset with prompts and descriptions introduced in \cite{Ngu_Towards_MICCAI2024}. As shown in \autoref{tab:text}, the reports are better to use than prompts, having significantly higher accuracy.
\begin{table}
\centering
\caption{Comparison of text classification accuracy on PCAM and BACH datasets using different types of textual inputs. We compare the prompts used in CONCH, and TQx prompt of \cite{Ngu_Towards_MICCAI2024} with the structured reports used in \clipit.}
\label{tab:text}
\begin{tabular}{|c|c|c|}
\hline
\textbf{Method}           & \textbf{PCAM} & \textbf{BACH} \\ \hline
Conch Prompt              & 62.99         & 36.26         \\ \hline
TQx         & 64.91         & 43.06         \\ \hline 
\clipit Text              & \textbf{70.29} & \textbf{55.50} \\ \hline
\end{tabular}
\end{table}

\subsection{Feature Visualization}
\begin{figure}[H]
    \centering
    \subfloat[\centering Without LoRA]{{\includegraphics[width=0.2\textwidth]{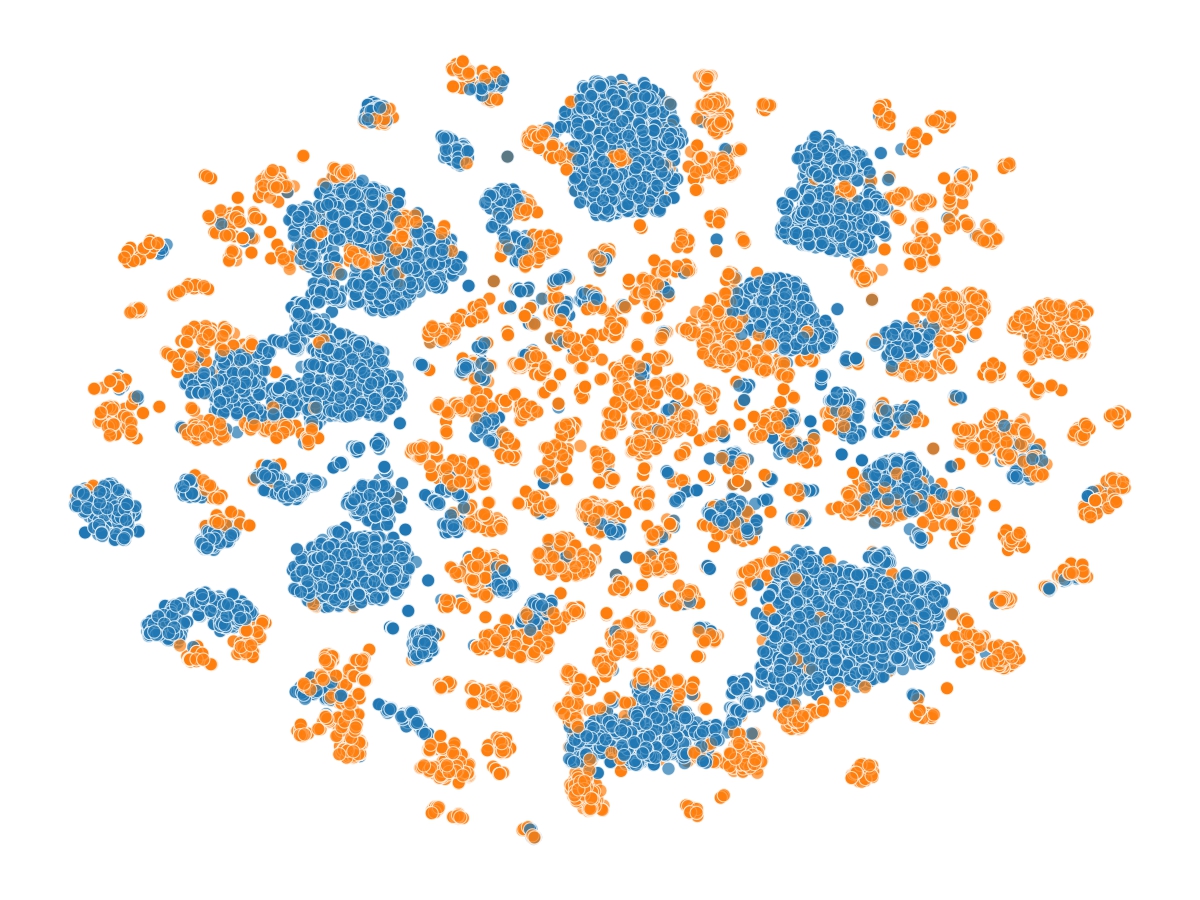} }}%
    \subfloat[\centering With LoRA]{{\includegraphics[width=0.2\textwidth]{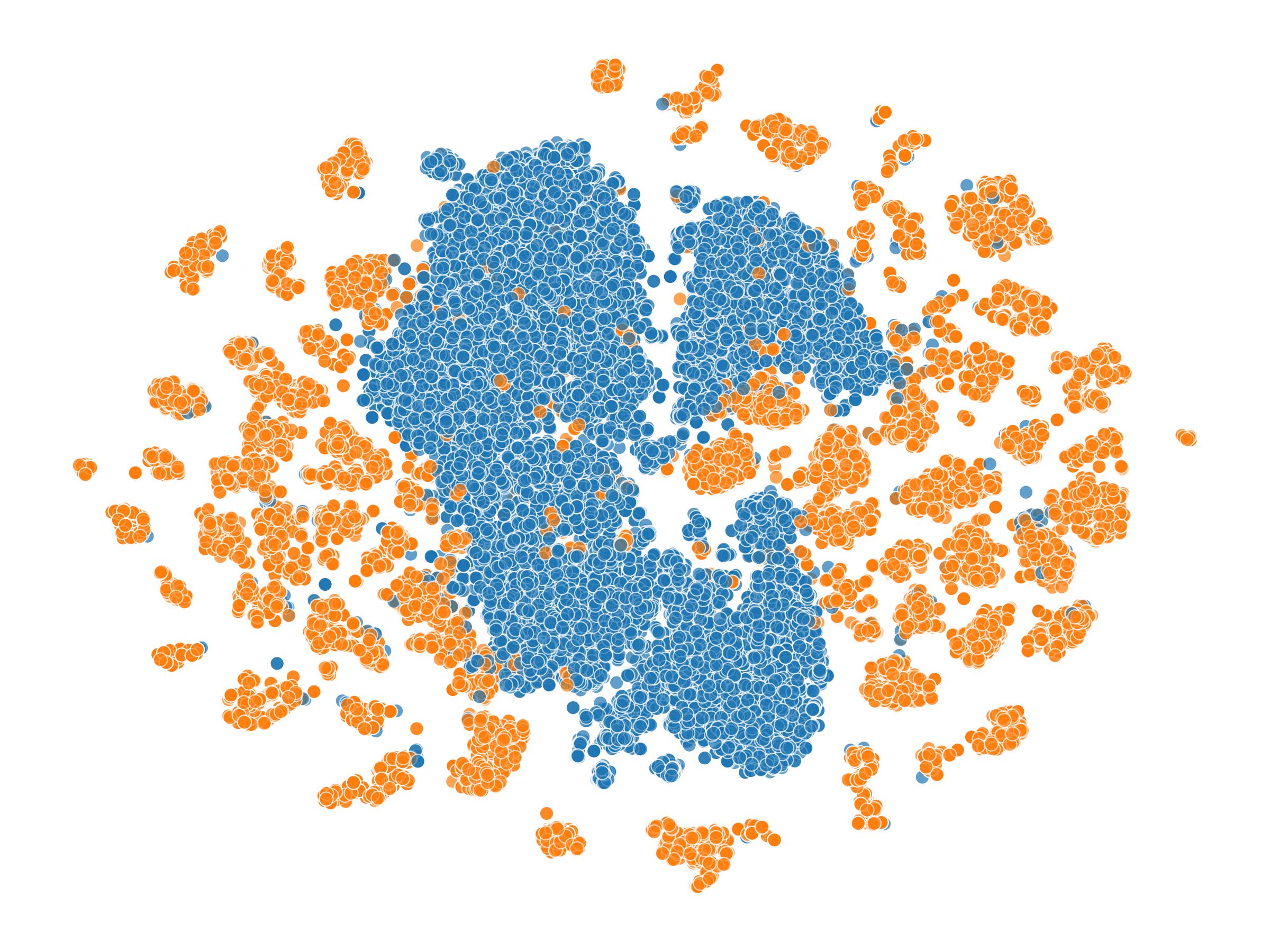} }}%
\caption{t-SNE plot of text-image embeddings, (a) without and (b) with LoRA for PCAM (blue: normal and yellow: tumor). This demonstrates LoRA's importance in aligning the image and text feature spaces, as improved class separability in the fused embedding space reflects better multimodal representation learning. } 
\label{fig:loraTSNE}
\end{figure}

To further analyze this effect, we visualize the joint text-image embeddings using t-SNE for the UNI model on PCAM, with and without LoRA. As shown in \autoref{fig:loraTSNE}, the embeddings without LoRA remain more scattered and class-overlapping, indicating poor class separability. In contrast, the embeddings after LoRA fine-tuning form distinct and well-separated clusters based on the class labels. This demonstrates that since the text and image embeddings are not fully aligned, not using LoRA will add noise to the fusion process.

\begin{figure}
    \centering
    \subfloat[\centering Text No LoRA]{{\includegraphics[width=0.2\textwidth]{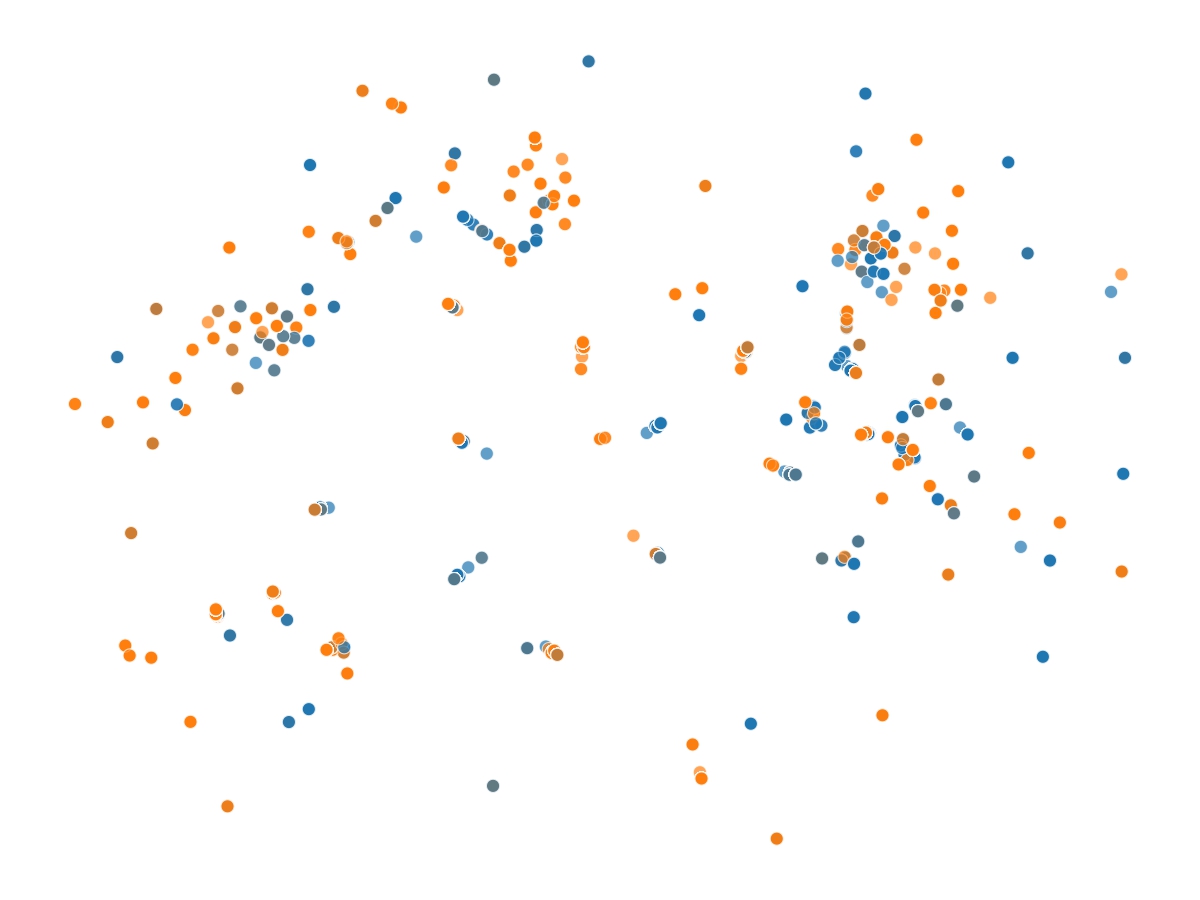} }}%
    \subfloat[\centering Text With LoRA]{{\includegraphics[width=0.2\textwidth]{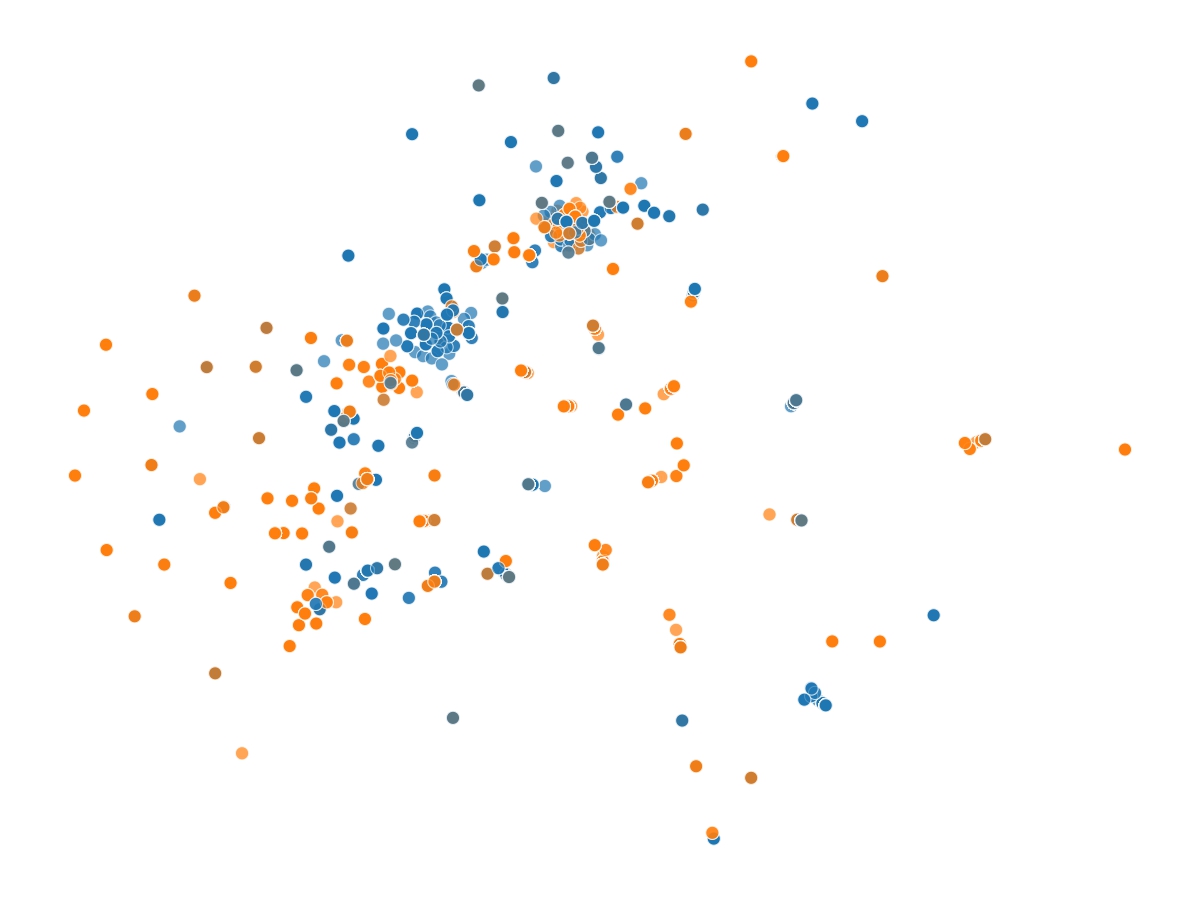} }}%
\caption{t-SNE plot of text embeddings with and without LoRA, for UNI-PCAM (blue: normal and yellow: tumor).} 
\label{fig:loraTSNEText}
\end{figure}

In order to further evaluate the influence of LoRA, we plot the t-SNE of text embeddings, with and without LoRA, as shown in \autoref{fig:loraTSNEText}. The t-SNE shows that the classes form significantly more compact and separable clusters.

\subsection{Text Statistics}
To better understand the performance differences of \clipit across datasets, we analyze the distribution of cosine similarity scores between the paired image and text embeddings for each dataset. \autoref{fig:sim} shows the similarity histograms for PCAM, BACH, and CRC after the modality pairing step using the CLIP-based model. As shown in the figure, both PCAM and BACH have approximately Gaussian-like distributions centered around a mean cosine similarity of near 0.55. This suggests that for these datasets, the CLIP model successfully identifies semantically relevant textual descriptions for most patches, leading to high-quality pairings. In contrast, the distribution for CRC is broader and skewed toward lower values, with a mean around 0.40. This indicates that retrieved reports for CRC patches show lower average semantic alignment, explaining smaller downstream gains. This observation aligns with the experimental results: \clipit yields the most significant performance improvements on PCAM and BACH, where the pairing quality is higher. For CRC, where the average similarity is lower, the benefit of using external reports as privileged information is reduced. 

\subsection{Distillation Ablation}
To evaluate the effectiveness of our representation-level distillation, we compute the cosine similarity between the ground-truth text embeddings and the predicted (distilled) text features generated from the image branch. As shown in Figure~\ref{fig:sim}, the distribution is sharply concentrated near 1, indicating high similarity between the original and distilled embeddings. This demonstrates that the text representations are accurately learned from the visual features, confirming the success of the distillation process in transferring semantic information from the text modality to the vision branch.

\begin{figure}[H]
    \centering
    {\includegraphics[width=0.30\textwidth]{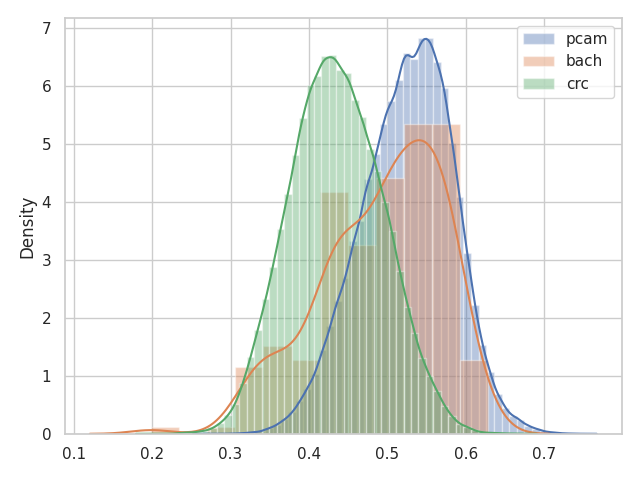} }
\caption{Histogram of cosine similarity scores between each histology image and its paired external report, as computed by the CLIP-based model during the modality pairing step. Distributions are shown for PCAM, BACH, and CRC datasets.} 
\label{fig:sim}
\end{figure}
\begin{figure}[H]
    \centering
    {\includegraphics[width=0.3\textwidth]{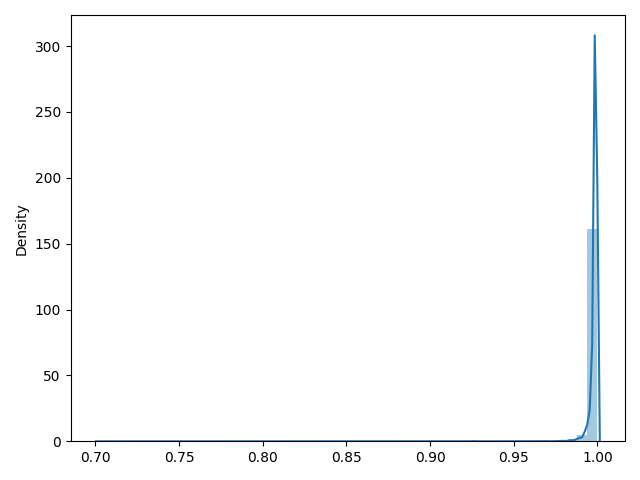} }
\caption{Histogram of cosine similarity between real text embedding and the distilled ones, showing that the text embedding was distilled perfectly to the vision.} 
\label{fig:sim}
\end{figure}
\subsection{Retrieval Backbone and Text Source}
To further assess the flexibility of \clipit, we ablate both the retrieval backbone and the source of textual data, as summarized in Table~\ref{tab1:ablation}. Specifically, we compare three CLIP-style retrieval models: (i) a generic CLIP model trained on natural images, (ii) QUILTNet, and (iii) CONCH, which is domain-pretrained on histology image–text pairs. In addition, we test two different textual resources: TCGA pathology reports and synthetic text prompts generated from class labels.
Results show that histology-pretrained retrieval backbones (e.g., CONCH, QUILT) consistently outperform generic CLIP, yielding gains of $1$–$2\%$ across backbones. This highlights the importance of domain-specific retrieval for maximizing pseudo-pair quality. However, even generic CLIP achieves measurable improvements over unimodal baselines, confirming that \clipit can still leverage external text supervision when only non-specialized retrieval models are available.

To further examine the dependency on specific text sources, we tested CLIP-IT with synthetic reports generated in a TCGA-like style. Specifically, we provided 10 real TCGA reports per dataset as in-context examples to a large language model (ChatGPT) and generated 500 synthetic reports for each class label. These synthetic reports were then used in place of the real TCGA reports during training. Using synthetic text in place of real reports yields performance comparable to TCGA reports in some cases (e.g., UNI, ViT-S/16), but generally underperforms real clinical text, underscoring the value of semantically rich, naturally occurring reports. 

Overall, these results demonstrate that \clipit is robust across retrieval backbones and text sources: domain-specific models and authentic reports provide the strongest benefits, but the framework remains effective even under weaker supervision.

\begin{table}[h!]
\centering
\vspace{-0.8em}
\caption{Ablation for CLIP model and text source.}
\resizebox{.99\linewidth}{!}{%
\begin{tabular}{l|cccc|c}
\hline
\label{tab1:ablation}
 \textbf{Text data} & \multicolumn{4}{c|}{\textbf{TCGA}} & \textbf{Synthetic} \\
\hline
\textbf{Model} & Unimodal & CLIP & Quilt & CONCH & CONCH \\
\hline
UNI     & 94.24 & 94.93 & 95.10 & 95.49 & 95.39 \\
DINO    & 88.88 & 90.35 & 91.34 & 92.32 & 89.91 \\
VITS16  & 88.13 & 90.14 & 91.09 & 91.42 & 90.85 \\
VITB16  & 88.43 & 89.82 & 92.84 & 90.93 & 90.14 \\
\hline
\end{tabular}
}
\end{table}
\subsection{Extension to Survival Prediction}
\label{Apx:survival}
While our main experiments focus on histology image classification, we emphasize that \clipit is task-agnostic by design. The pseudo-pairing and distillation pipeline can be integrated with a variety of downstream objectives beyond classification. In response to reviewer feedback, we conducted a preliminary experiment on survival prediction using the Metastatic Breast Cancer (MBC) whole-slide image dataset~\cite{Bergstrom2024, Galland2022}. We employed UNI as the patch-level feature extractor and AttentionMIL~\cite{ilse2018attention} with a Cox head for survival analysis. 

Across five folds, the multimodal \clipit approaches improved survival prediction compared to the unimodal baseline, raising the mean C-index from $0.609$ to $0.661$ ($+0.052$) and the mean iAUC from $0.633$ to $0.699$ ($+0.066$) (~\autoref{tab:survival}). Improvements were consistent across four folds, with only one fold showing a slight decrease. These results indicate that incorporating textual embeddings provides measurable gains in survival discrimination. 
\begin{table}[h!]
\centering
\caption{Survival prediction results on the MBC dataset.}
\label{tab:survival}
\renewcommand{\arraystretch}{1.2}
\resizebox{0.9\linewidth}{!}{%
\begin{tabular}{l|cc|cc}
\toprule
\multirow{2}{*}{\textbf{Fold}} & \multicolumn{2}{c|}{\textbf{C-index}} & \multicolumn{2}{c}{\textbf{iAUC}} \\
& Unimodal & \clipit & Unimodal & \clipit \\
\midrule
\textbf{Mean} & 0.609 & \textbf{0.661} & 0.634 & \textbf{0.699} \\
\textbf{$\Delta$} & \multicolumn{2}{c}{\textbf{+0.052}} & \multicolumn{2}{c}{\textbf{+0.066}}  \\
\bottomrule
\end{tabular}
}
\end{table}
Overall, these findings support our claim that \clipit generalizes beyond image classification and can enhance performance in more complex tasks such as survival analysis.